\documentclass[10pt,twocolumn,letterpaper]{article}

\usepackage[pagenumbers]{cvpr} 
\usepackage[accsupp]{axessibility}  

\makeatletter
\@namedef{ver@everyshi.sty}{}
\makeatother
\usepackage{tikz}
\usepackage{graphicx}
\usepackage{amsmath}
\usepackage{amssymb}
\usepackage{booktabs}

\usepackage{tikz}
\usepackage{comment}
\usepackage{color}
\usepackage{svg}
\usepackage{subcaption}
\usepackage{algorithm}
\usepackage{algpseudocode}

\usepackage[pagebackref,breaklinks,colorlinks]{hyperref}

\usepackage[capitalize]{cleveref}
\crefname{section}{Sec.}{Secs.}
\Crefname{section}{Section}{Sections}
\Crefname{table}{Table}{Tables}
\crefname{table}{Tab.}{Tabs.}

\def\cvprPaperID{7583} 
\def\confName{CVPR}
\def\confYear{2023}

\graphicspath{{Figures/}}

\newif\ifdraft
\draftfalse

\newcommand{\x}{\mathbf{x}}

\newcommand{\y}{\mathbf{y}}

\newcommand{\T}{\mathcal{T}}

\newcommand{\M}{\mathcal{M}}

\DeclareMathOperator*{\argmin}{arg\,min}

\algrenewcommand\algorithmicrequire{\textbf{Input:}}
\algrenewcommand\algorithmicensure{\textbf{Output:}}
\algdef{SE}{Repeat}{EndRepeat}[1]{\textbf{repeat} \(\mbox{#1}\) \textbf{times}}{\textbf{end repeat}}%

\begin{document}

\title{Full or Weak annotations? \\An adaptive strategy for budget-constrained annotation campaigns}


\author{Javier Gamazo Tejero$^1$, Martin S. Zinkernagel$^2$, Sebastian Wolf$^2$\\
Raphael Sznitman$^1$, Pablo Márquez Neila$^1$\\
{\small     $^1$University of Bern, $^2$Inselspital Bern, Switzerland}\\
{\tt\small \{javier.gamazo-tejero, raphael.sznitman, pablo.marquez\}@unibe.ch}\\
{\tt\small \{martin.zinkernagel, sebastian.wolf\}@insel.ch}
}

\maketitle

\begin{abstract}
Annotating new datasets for machine learning tasks is tedious, time-consuming, and costly. For segmentation applications, the burden is particularly high as manual delineations of relevant image content are often extremely expensive or can only be done by experts with domain-specific knowledge. Thanks to developments in transfer learning and training with weak supervision, segmentation models can now also greatly benefit from annotations of different kinds. However, for any new domain application looking to use weak supervision, the dataset builder still needs to define a strategy to distribute full segmentation and other weak annotations. Doing so is challenging, however, as it is a priori unknown how to distribute an annotation budget for a given new dataset. To this end, we propose a novel approach to determine annotation strategies for segmentation datasets, whereby estimating what proportion of segmentation and classification annotations should be collected given a fixed budget. To do so, our method sequentially determines proportions of segmentation and classification annotations to collect for budget-fractions by modeling the expected improvement of the final segmentation model. We show in our experiments that our approach yields annotations that perform very close to the optimal for a number of different annotation budgets and datasets.

\end{abstract}


\section{Introduction}
\label{sec:intro}
Semantic segmentation is a fundamental computer vision task with applications in numerous domains such as autonomous driving~\cite{cordts2016cityscapes,siam2017deep}, scene understanding~\cite{sless2019road}, surveillance~\cite{tseng2021person} and medical diagnosis~\cite{chen2020deep,hesamian2019deep}. As the advent of deep learning has significantly advanced the state-of-the-art, many new application areas have come to light and continue to do so too. This growth has brought and continues to bring exciting domain-specific datasets for segmentation tasks~\cite{islam2020semantic,li2020mas3k,Bodenstedt2018,liu2020fsd,WelinderEtal2010}. 

Today, the process of establishing machine learning-based segmentation models for any new application is relatively well understood and standard. Only once an image dataset is gathered and curated, can machine learning models be trained and validated. In contrast, building appropriate datasets is known to be difficult, time-consuming, and yet paramount. Beyond the fact that collecting images can be tedious, a far more challenging task is producing ground-truth segmentation annotations to subsequently train  (semi) supervised machine learning models. This is mainly because producing segmentation annotations often remains a manual task. As reported in~\cite{Bearman16}, generating segmentation annotations for a single PASCAL image~\cite{pascal-voc-2012} takes over 200 seconds on average. This implies over 250 hours of annotation time for a dataset containing a modest 5'000 images. What often further exacerbates the problem for domain-specific datasets is that only the dataset designer, or a small group of individuals, have enough expertise to produce the annotations (\eg, doctors, experts, etc.), making crowd-sourcing ill-suited. 
\begin{figure*}[t]
\centering
\includegraphics[width=0.99\textwidth]{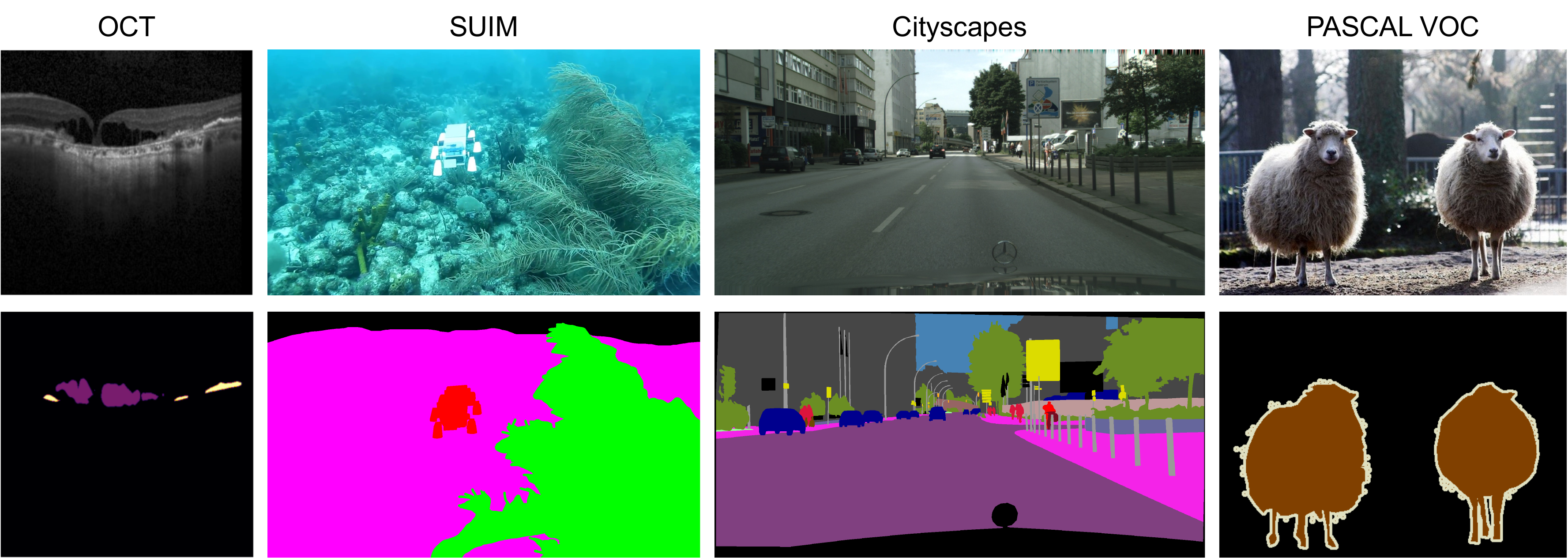}
\caption{Illustration of different semantic segmentation applications; OCT: Pathologies of the eye in OCT images, SUIM: Underwater scene segmentation~\cite{islam2020semantic}, Cityscape: street level scene segmentation~\cite{cordts2016cityscapes}, PASCAL VOC: natural object segmentation.}
\label{fig:datasets}
\end{figure*}

To overcome this challenge, different paradigms have been suggested over the years. Approaches such as Active Learning~\cite{Cai21,Casanova2020Reinforced,Konyushkova15} aim to iteratively identify subsets of images to annotate so as to yield highly performing models. Transfer learning has also proved to be an important tool in reducing annotation tasks~\cite{Ding2019,heker2020joint,kolesnikov2020big,koleshnikov2021,Liang2020,menegola2017}. For instance, \cite{Mensink}~show that training segmentation models from scratch is often inferior to using pre-training models derived from large image classification datasets, even when the target application domain differs from the source domain. Finally, weakly-supervised methods~\cite{ahn2018learning,Papandreou15} combine pixel-wise annotations with other weak annotations that are faster to acquire, thereby reducing the annotation burden. In particular, Papandreou~\etal~\cite{Papandreou15} showed that combinations of strong and weak annotations (\eg, bounding boxes, keypoints, or image-level tags) delivered competitive results with a reduced annotation effort. In this work, we rely on these observations and focus on the weakly supervised segmentation setting.

In the frame of designing annotation campaigns, weakly-supervised approaches present opportunities for efficiency as well. Instead of completely spending a budget on a few expensive annotations, weakly-supervised methods allow a proportion of the budget to be allocated to inexpensive, or weak, labels. That is, one could spend the entire annotation budget to manually segment available images, but would ultimately lead to relatively few annotations. Conversely, weak annotations such as image-level labels are roughly 100~times cheaper to gather than their segmentation counterparts~\cite{Bearman16}. Thus, a greater number of weakly-annotated images could be used to train segmentation models at an equal cost. In fact, under a fixed budget, allocating a proportion of the budget to inexpensive image-level class labels has been shown to yield superior performance compared to entirely allocating a budget to segmentation labels~\cite{Bearman16}.

Yet, allocating how an annotation budget should be distributed among strong and weak annotations is challenging, and inappropriate allocations may severely impact the quality of the final segmentation model. For example, spending the entire budget on image-level annotations will clearly hurt the performance of a subsequent segmentation model. Instead, a naive solution would be to segment and classify a fixed proportion of each (\eg,~say 80\% - 20\%). Knowing what proportion to use for a given dataset is unclear, however. Beyond this, there is no reason why the same fixed proportion would be appropriate across different datasets or application domains. That is, it would be highly unlikely that the datasets shown in Fig.~\ref{fig:datasets} all require the same proportion of strong and weak annotations to yield optimal segmentation models.

Despite its importance, choosing the best proportion of annotation types remains a largely unexplored research question. Weakly-supervised and transfer-learning methods generally assume that the annotation campaign and the model training are independent and that all annotations are simply available at training time. While active learning methods do alternate between annotation and training, they focus on choosing optimal samples to annotate rather than choosing the right type of annotations. Moreover, most active learning methods ignore constraints imposed by an annotation budget. More notable, however, is the recent work of Mahmood {\it et.~al.}~\cite{mahmood2022, mahmood2022optimizing} which aims to determine what weak and strong annotation strategy is necessary to achieve a target performance level. While noteworthy, this objective differs from that here, whereby given a fixed budget, what strategy is best suited for a given new dataset?

To this end, we propose a novel method to find an optimal budget allocation strategy in an online manner. Using a collection of unlabeled images and a maximum budget, our approach selects strong and weak annotations, constrained by a given budget, that maximize the performance of the subsequent trained segmentation model. To do this, our method iteratively alternates between partial budget allocations, label acquisition, and model training. At each step, we use the annotations performed so far to train multiple models to estimate how different proportions of weak and strong annotations affect model performance. A Gaussian Process models these results and maps the number of weak and strong annotations to the expected model improvement. Computing the Pareto optima between expected improvement and costs, we choose a new sub-budget installment and its associated allocation so to yield the maximum expected improvement. We show in our experiments that our approach is beneficial for a broad range of datasets, and illustrate that our dynamic strategy allows for high performances, close to optimal fixed strategies that cannot be determined beforehand.

\section{Related work}
\label{sec:related}

\subsection{Weak annotations for segmentation}
Weakly supervised semantic segmentation (WSSS) relies on coarser annotations, such as bounding boxes~\cite{song2019box}, scribbles~\cite{lin2016scribblesup,tang2018normalized} or image-level classification labels~\cite{ahn2019weakly}, to train a segmentation network. WSSS methods have often employed saliency maps as weak annotations for segmentation models, as these are typically obtained from CAM~\cite{zhou2016learning}, which leverages image-level classification annotation. These methods then focus on refining the saliency maps with a variety of techniques~\cite{fan2020learning,lee2019ficklenet}. Others make use of attention to achieve coarse segmentations~\cite{jiang2019integral,Ki2021}. Conversely, \cite{Zhang2021} combined annotations in the form of bounding boxes and image-level labels to accurately generate image graphs, to be used by a graph neural network to predict node values corresponding to pixel labels. In this context, the work in~\cite{mahmood2022} and~\cite{mahmood2022optimizing} are close to this one, whereby their objective is to determine what annotation strategy over annotation types is likely to yield a target performance level. 

\subsection{Transfer learning}
Due to the limited availability of annotated image data in some domains, it is now common to use neural networks pre-trained on large image classification tasks~\cite{deng2009imagenet} for subsequent target tasks. Specifically, in cases where the target task has limited data or annotations, this has been shown to be particularly advantageous. Among others, this practice is now widely used in medical imaging and has been linked to important performance gains after fine-tuning~\cite{Ding2019,Esteva2017,menegola2017,Tajbakhsh2016,kolesnikov2020big}.


Efforts are now pivoting towards the use of in-domain pre-training, avoiding the leap of faith that is often taken with Imagenet~\cite{heker2020joint,Liang2020}. In~\cite{Liang2020}, the model is pre-trained on ChestX-ray14~\cite{wang2017chestxray} to more accurately detect pneumonia in chest X-ray images from children. In~\cite{heker2020joint}, the authors show that joint classification and segmentation training, along with pre-training on other medical datasets that have domain similarity, increases segmentation performances with respect to the segmentation using Imagenet-based pre-training.

Alternatively, cross-task methods seek to transfer features learned on one task (\eg~ classification, normal estimation, etc.) to another, usually more complex one. Along this line, Taskonomy~\cite{Zamir2018} explored transfer learning capabilities among a number of semantic tasks and built a task similarity tree that provided a clustered view of how much information is available when transferring to other tasks. Similarly,~\cite{Mensink} performed an extensive study of cross-task transfer capabilities for a variety of datasets, reaching the conclusion that Imagenet pre-training outperforms random initialization in all cases, but further training on related tasks or domains also brings additional benefits.



\subsection{Active learning}
In active learning, the goal is to train a model while querying an oracle to label new samples that are expected to improve the model's accuracy. In computer vision, it has been applied to image classification~\cite{Joshi2009,Ranganathan2017} or semantic segmentation~\cite{andriluka2018fluid,Benenson2019LargeScaleIO,siddiqui2020viewal} among others. As a byproduct, Active learning has also been used as a way to reduce labeling time. For example,~\cite{Konyushkova_2018_CVPR} describes a method that couples Reinforcement Learning and Active Learning to derive the shortest sequence of annotation actions that will lead to object detection within an image. Others have focused on speeding up this process via eye-tracking~\cite{papadopoulos2014} or extreme clicking~\cite{papadopoulos2017extreme}. As such, Active Learning is related to the present work in the sense that our approach is adaptive but differs in that our method determines what annotations types should be collected under a constrained budget instead of predicting at each time step which samples should be added to the annotated set.


\section{Method}
\label{sec:method}


Training segmentation models using a combination of expensive pixel-wise annotations and other types of cheaper annotations, such as image-wise labels or single-pixel annotations is known to be beneficial, as well as using cross-task transfer learning techniques~\cite{Mensink}. This is motivated by empirical findings showing that, under a limited annotation budget, allocating a proportion of the budget to inexpensive image-level class labels led to superior performance compared to allocating the budget entirely to segmentation labels~\cite{Bearman16}. However, the optimal proportion of the budget to allocate per annotation type is a-priori unknown beforehand and data-dependent. Thus, the goal of our method is to find this data-specific optimal budget allocation in an online manner, as it is necessary for any dataset builder starting off.

We describe our method in the subsequent sections. For clarity, we focus on image segmentation and assume two kinds of annotations are possible: strong annotations as segmentation labels and weak annotations as image-level classification labels. Generalizing this formulation to other tasks or settings with more than two annotations types should follow directly.

\subsection{Problem formulation} 
Let $p_\textrm{data}(\x)$ be the distribution of training images for which we have no annotations initially. Each training image~$\x$ can be annotated with a pixel-wise segmentation labeling~{$(\x,\y)\sim{}p_\textrm{data}(\x)p_\textrm{sgm}(\y\mid\x)$} or an image-wise classification annotation~{$(\x,c)\sim{}p_\textrm{data}(\x)p_\textrm{cls}(c\mid\x)$}
Sampling from the distributions $p_\textrm{cls}$ and~$p_\textrm{sgm}$ represents the task of manually annotating the image and has associated costs of~$\alpha_\textrm{c}>0$ and ~$\alpha_\textrm{s}>0$, respectively. Supported by previous work~\cite{Bearman16,Mensink,mahmood2022}, we will assume that $\alpha_\textrm{s}\gg\alpha_\textrm{c}$.

By sampling $C$~classifications from $p_\textrm{cls}$ and $S$~segmentation from ~$p_\textrm{sgm}$, we can build an annotated training dataset $\T=(\T_c,\T_s)\sim{}(p_\textrm{cls}^C, p_\textrm{sgm}^S)$. The dataset~$\T$ then has an annotation cost,
\begin{equation}
    \alpha_\textrm{c}C+\alpha_\textrm{s}S,
\end{equation}
which we assume to be bounded by an upper limit, or \emph{budget},~$B$.

To annotate $\T$, however, we can choose different \emph{allocation strategies}, or combinations of $C$ and~$S$, that have different costs and that yield different segmentation model performances. The utility~$u$ of an allocation strategy~$(C, S)$ is the expected performance of a model trained with datasets that follow that strategy,
\begin{equation}
    \label{eq:utility}
    u(C, S) = \mathbb{E}_{(\T_c,\T_s) \sim{}(p_\textrm{cls}^C, p_\textrm{sgm}^S)} \left[m(\T_c, \T_s)\right],
\end{equation}
\noindent
where~$m(\T_c, \T_s)$ is the performance score (\eg,~Dice score, IoU) of a segmentation model trained with datasets~($\T_c, \T_s$) and evaluated on a separate fixed test dataset. Note that in contrast to Active Learning, the utility is defined over the set of strategies~$(C,S)$ and not over the individual samples of a fixed training set. This is motivated by our aim to estimate the performance of the annotation strategy~$(C, S)$ and not the ensuing specific training dataset. 

\begin{figure*}[h]
\centering
\includegraphics[width=.95\textwidth]{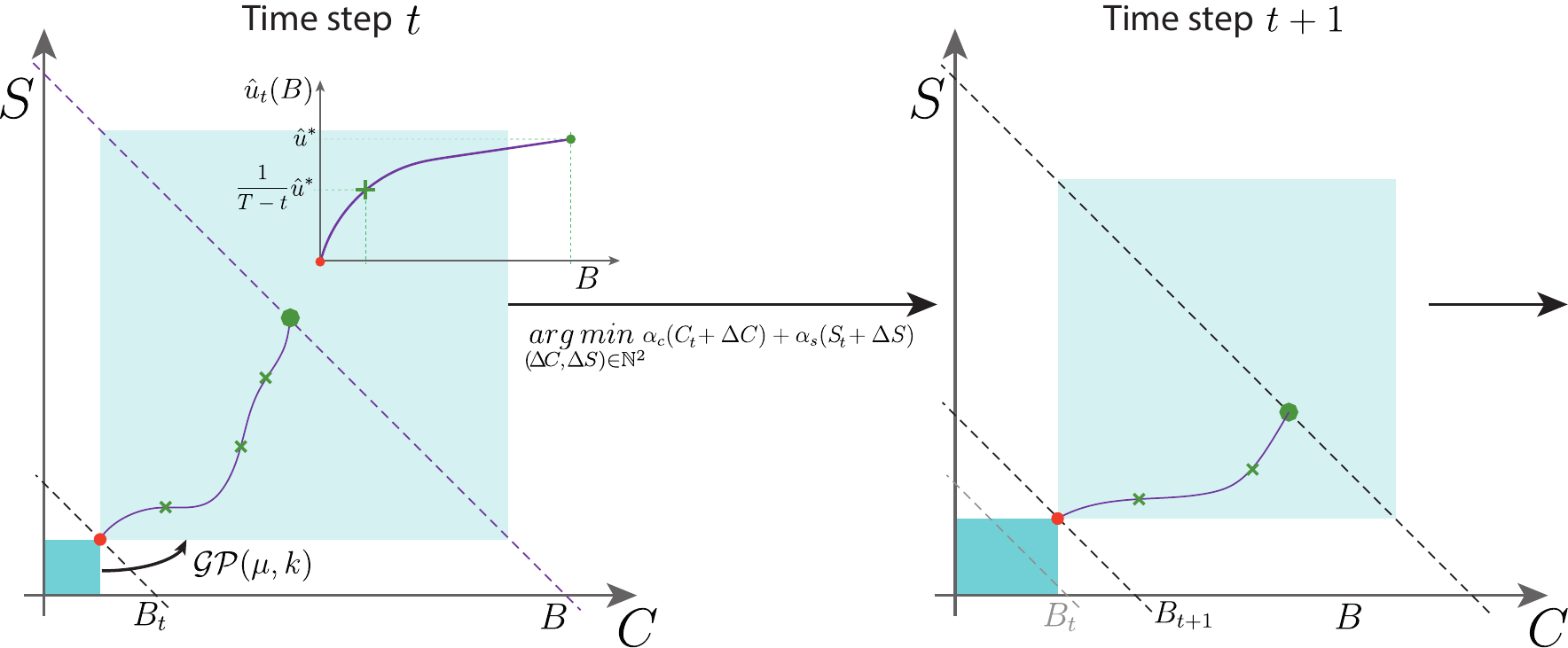}
\caption{Illustration of proposed method. At a given iteration $t$, $C_t$ and $S_t$ classification and segmentation annotations have already been collected (blue region, left panel) with a budget of $B_t$. For the next annotation phase, the budget is increased to $B_{t+1}$. To determine how many new classification and segmentation annotations to collect, $M$ combinations of different quantities $(C^{(i)}, S^{(i)})$ are gathered according to Alg.~\ref{alg:build_gp_samples} to compute $m(C^{(i)}, S^{(i)})$. A Gaussian Process is then trained to estimate the utility of different combinations of annotation types (light blue area, left panel). From this, we infer $\Delta C$ and $\Delta S$ to select next by computing the combination that maximizes the expected improvement along the Pareto front given by the budget $B_2$ (red point, left panel). The next iteration starts then with the new proportions (red point, right panel) and follows the same steps (see text and Alg.~\ref{alg:the_algorithm} for details). For illustration purposes, the costs are set here to $\alpha_c=\alpha_s=1$.
}
\label{fig:method}
\end{figure*}

Our goal then is to find the annotation strategy that maximizes the expected performance constrained to a budget~$B$,
\begin{equation}
\label{eq:the_optimization}
\begin{aligned}
    \max_{(C,S)\in\mathbb{N}^2} \quad & u(C, S), \\
    \textrm{s.t.} \quad & \alpha_cC+\alpha_sS \le B.
\end{aligned}
\end{equation}
In the following, we describe how we optimize Eq.~\eqref{eq:the_optimization}.


\subsection{Utility model}

As defined Eq.~\eqref{eq:utility}, the utility function,~$u$, marginalizes over all possible training sets, which is intractable to compute in practice. To overcome this computational challenge, we approximate~$u$ with a collection~$\M$ of discrete samples, where each sample $m\in\M$ is a tuple containing an allocation strategy~$(C, S)$ and the estimated score $m(\T_c, \T_s)$ obtained for a dataset sampled with that allocation strategy. To build~$\M$, one could simply sample a random strategy~$(C', S')$, annotate a dataset~$(\T'_c, \T'_s)\sim{}(p_\textrm{cls}^{C'}, p_\textrm{sgm}^{S'})$, and measure its performance. However, this would imply annotating for different potential budgets and is thus infeasible in practice. Instead, a practical alternative is to leverage previously annotated data~$(\T_c, \T_s)$. For each sampled strategy~$(C', S')$, we build the corresponding dataset~$(\T'_c, \T'_s)$ by taking random samples from the already annotated data according to the strategy. While this procedure, formalized in Alg.~\ref{alg:build_gp_samples}, leads to biased samples, we empirically found this bias to have a minor impact on the final strategies compared to estimations with unbiased sampling.

\begin{algorithm}[t!] 
\caption{Build utility samples from annotated data}
\label{alg:build_gp_samples}
\begin{algorithmic}[1]
    \Function{BuildUtilitySamples}{$\T_c,\T_s$}
    \State $C \leftarrow |\T_c|$, $S \leftarrow |\T_s|$
    \State $\mathcal{M} \leftarrow \{((C, S), m(\T_c,\T_s))\}$ \Comment{Add sample with all the available data}
    \Repeat {$M-1$}
    \State Sample $(C', S')\in[0, C]\times[0, S]$
    \State $\T'_c \leftarrow$ \{$C'$~elements sampled from $\T_c$\}
    \State $\T'_s \leftarrow$ \{$S'$~elements sampled from $\T_s$\}
    \State $\mathcal{M} \leftarrow \mathcal{M} \cup ((C', S'), m(\T'_c, \T'_s))$ 
    \EndRepeat
    \EndFunction
\end{algorithmic}
\end{algorithm}



While $\M$~provides an estimation of~$u$ as a set of discrete locations, we generalize these estimations to the entire space of strategies by fitting a Gaussian process~(GP) to the samples in~$\M$. The Gaussian process, $\mathcal{GP}(\mu, k)$ is parameterized by a suitable mean function~$\mu$ and covariance function~$k$. 
In our case, we use the mean function,
\begin{equation}
    \mu(C, S) = \gamma_c\log(\beta_cC+1) + \gamma_s\log(\beta_sS+1),
    \label{eq:gp_mean}
\end{equation}
which accounts for the fact that the segmentation performance increases logarithmically with the volume of the training data~\cite{sun2017} and that each annotation type has a different rate of performance growth. Similarly, the covariance~$k$ is a combination of two RBF~kernels with different scales~$\ell_c$, $\ell_s$ for each annotation type,
\begin{equation}
    k\left((C, S), (C', S')\right) = \sigma^2e^{-\frac{(C-C')^2}{2\ell_c^2}} e^{-\frac{(S-S')^2}{2\ell_s^2}}.
\end{equation}
The values~$\gamma_c$, $\beta_c$, $\gamma_s$, $\beta_s$ from the mean, the length scales~$\ell_c$, $\ell_s$ and the amplitude~$\sigma$ from the covariance are trainable parameters of the~GP.

The trained GP models a distribution over utility functions, $u \sim\mathcal{GP}(\mu, k)$, that are plausible under the samples~$\M$. This distribution represents not only the expected utility, but also its uncertainty in different areas of the strategy space. Sampling just a single~$u$ from the GP to solve Eq.~\eqref{eq:the_optimization} would thus be suboptimal. For this reason, we substitute the utility~$u$ in Eq.~\eqref{eq:the_optimization} by a surrogate function~$\hat{u}$ that trades-off exploitation and exploration, thus incorporating uncertainty information into the optimization problem. Following a Bayesian optimization approach~\cite{jones1998efficient}, we choose~$\hat{u}$ to be the expected improvement~(EI),
\begin{equation}
    \hat{u}(C, S) = \mathbb{E}_{u\sim\mathcal{GP}_t}[\max\{u(C, S) - m^*, 0\}],
    \label{eq:EI}
\end{equation}
where~$m^*$ is the current maximum point.


\subsection{Optimization}


Training the GP requires annotated data to build the set~$\M$, which in turn relies on an annotation strategy that we are trying to find, whereby implying a circular dependency. We address this circular dependency by optimizing Eq.~\eqref{eq:the_optimization} in an iterative manner.

Our algorithm shown in Alg.~\ref{alg:the_algorithm}, allocates the available budget~$B$ in a fixed number of adaptive installments, alternating between data annotation with the current strategy, GP~fitting, and strategy selection for the next budget installment. More specifically, our method starts with an initial strategy~$(C_0, S_0)$ with associated cost~$B_0$. At each iteration~$t$, new data is annotated according to the current strategy~$(C_t, S_t)$ so that the sets of annotated data~$(\T_c, \T_s)$ contain $C_t$~classification and $S_t$~segmentation annotations, respectively. From the available annotated data~$(\T_c, \T_s)$, we extract new samples for~$\M$ and fit the GP, which defines the surrogate function~$\hat{u}_t$. The corresponding current maximal point~$m^*_t$ is set to be the maximum performance found so far, (\ie, the performance of the model trained with all the annotated data available at this iteration), $m^*_t=m(\T_c, \T_s)$. Finally, this surrogate function is used to estimate the next best strategy~$(C_{t+1}, S_{t+1})$. We find a delta strategy~$(\Delta{}C,\Delta{}S)$ that increases the expected improvement by a fixed fraction of its maximum possible value,
\begin{equation}
\begin{aligned}
    \argmin_{(\Delta{}C,\Delta{}S)\in\mathbb{N}^2} \quad & \alpha_c (C_t + \Delta{}C)+\alpha_s (S_t + \Delta{}S), \\
    \textrm{s.t.} \quad & \hat{u}_t(C_t + \Delta{}C, S_t + \Delta{}S) \ge \dfrac{1}{T - t}\hat{u}^*_t,
\end{aligned}
\label{eq:delta_strategy}
\end{equation}

where $T$~is the desired maximum number of iterations of the algorithm and~$\hat{u}_t^*$ is the maximum expected improvement that can be reached using the entire budget~$B$ for the current surrogate function~$\hat{u}_t$ according to Eq.~\eqref{eq:the_optimization}. The found delta strategy defines the new strategy~$(C_{t+1}, S_{t+1}) = (C_t + \Delta{}C, S_t + \Delta{}S)$ for the next iteration. The process is depicted in Fig.~\ref{fig:method}.

\begin{algorithm}[b]
\caption{Proposed approach}
\label{alg:the_algorithm}
\begin{algorithmic}[1]
\Require Number of iterations~$T$, initial labelling strategy~$(C_0, S_0)$
\State $t\leftarrow 0$, $\Delta{}C\leftarrow C_0$, $\Delta{}S\leftarrow S_0$, $\T_c=\emptyset$, $\T_s=\emptyset$, $\mathcal{M}=\emptyset$
\While {$t<T$}
\State Annotate new data $(\Delta{}\T_c,\Delta{}\T_s)\sim{}(p_\textrm{cls}^{\Delta{}C}, p_\textrm{sgm}^{\Delta{}S})$
\State $\T_c \leftarrow \T_c \cup \Delta{}\T_c, \quad \T_s \leftarrow \T_s \cup \Delta{}\T_s$ \Comment{Note that $|\T_c|=C_t$ and $|\T_s|=S_t$}
\State $\mathcal{M} \leftarrow \mathcal{M} \,\cup\,$\Call{BuildUtilitySamples}{$\T_c$, $\T_s$} 
\State Train GP with samples in~$\mathcal{M}$
\State Compute $(\Delta{}C, \Delta{}S)$ from Eq.~\eqref{eq:delta_strategy}
\State $C_{t+1} \leftarrow C_t + \Delta{}C, \quad S_{t+1} \leftarrow S_t + \Delta{}S$
\State $t\leftarrow t+1$
\EndWhile
\Ensure $(C_T, S_T)$
\end{algorithmic}
\end{algorithm}

Note that solving Eq.~\eqref{eq:delta_strategy} requires finding~$\hat{u}_t^*$, which in turn requires solving Eq.~\eqref{eq:the_optimization}. While solving two optimization problems may seem unnecessary, the solutions of both problems are in the Pareto front of strategies (\ie, the set of non-dominated strategies for which no other strategy has simultaneously smaller cost and larger or equal expected improvement). Given that the space of strategies is discrete, the elements of the Pareto front can be easily found in linear time by enumerating all possible strategies, computing their costs and expected improvements with~$\hat{u}_t$, and discarding the dominated elements. Given the Pareto front, the strategy with the maximum EI~$u^*_t$ and the strategy of minimum budget with EI larger than $\frac{1}{T - t}\hat{u}^*_t$, which correspond to the solutions of Eq.\eqref{eq:the_optimization} and Eq.~\eqref{eq:delta_strategy}, respectively, can be found in linear time.

\section{Experimental setup}
\label{sec:experiments}

To validate our approach, we evaluated it on four different datasets, while comparing its performance to a set of typical fixed budget allocation strategies. In addition, we explore the impact of different hyper parameters on the overall performance of the method. 
\begin{figure*}[h]
\centering
\includegraphics[width=0.96\textwidth]{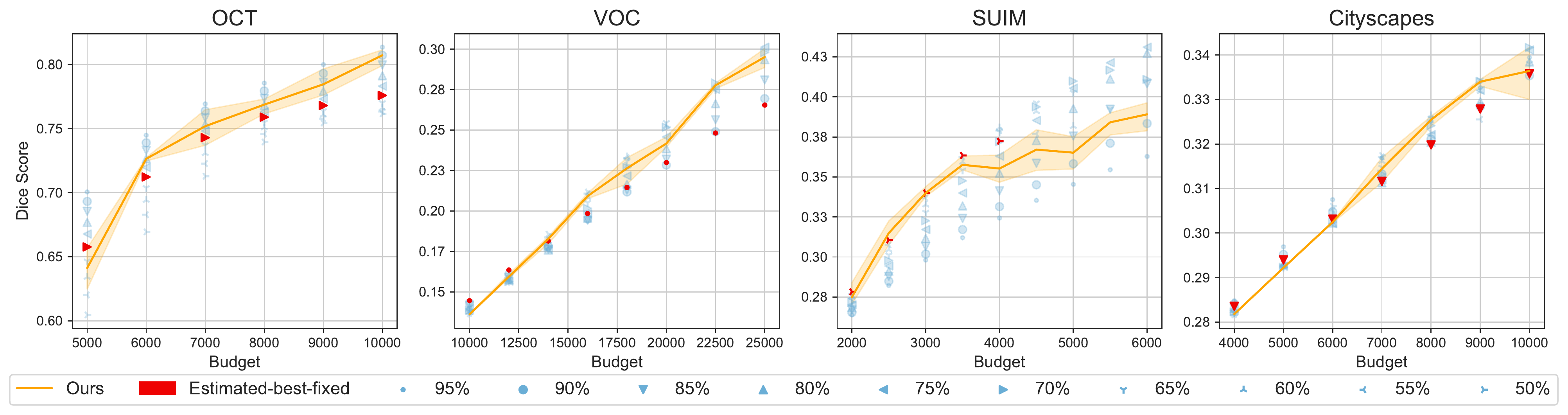}
\caption{
Performance of our method (orange line) on OCT, PASCAL VOC, SUIM and Cityscapes datasets. Shaded region is computed from three seeds. Fixed strategies are shown in blue. Red points show the {\it estimated-best-fixed} strategy with $B_0$.  Labels expressed as percentage of the budget allocated to segmentation. Note that the first budget $B$ fulfills $B\gg B_0$ in all cases. 
}
\label{fig:main_results}
\end{figure*}

\begin{figure*}[t]
\centering
\includegraphics[width=0.96\textwidth]{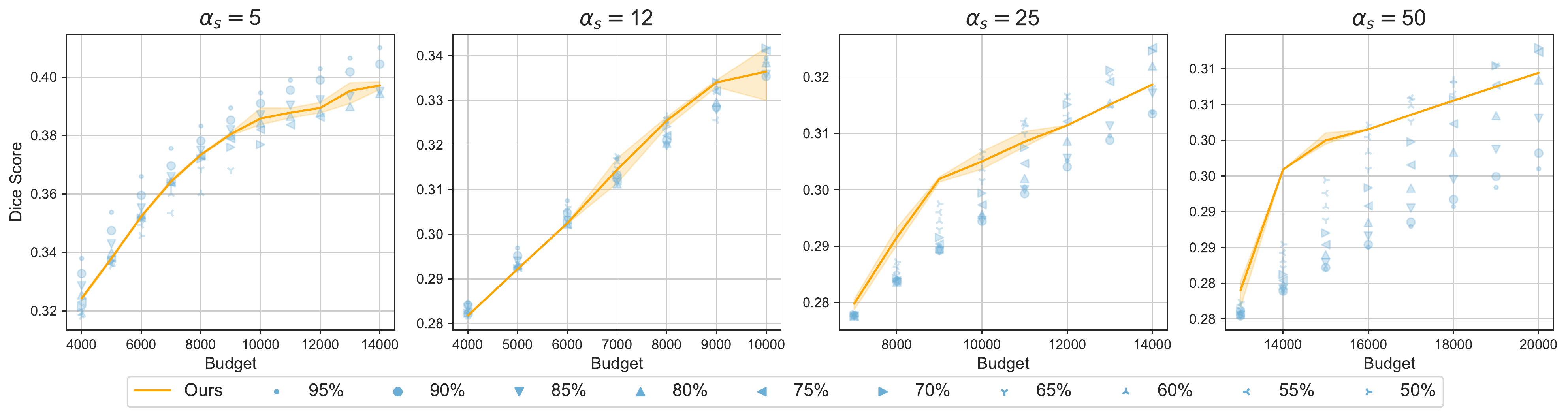}
\caption{Mean of our method with $\alpha_s = \{5, 12, 25, 50\}$ on Cityscapes (orange, line). Shaded region is computed from three seeds. Fixed strategies are shown in blue. Labels expressed as percentage of the budget allocated to segmentation.
}
\label{fig:alpha_results}
\end{figure*}

\subsection{Datasets} 
We chose a collection of datasets with different image modalities, including a medical dataset as they often suffer from data and annotation scarcity. In this context, they represent a typical new application domain where our method could be particularly helpful. In each case, we enumerate the number of images for which classification or segmentation images can be sampled by a method:
\begin{description}
    \item[Augmented PASCAL VOC 2012~\cite{pascal-voc-2012}:] 5'717~classification and 10'582~segmentation natural images with 21~classes for training. The validation sets contain 1'449 segmented images.
    \item[SUIM~\cite{islam2020semantic}:] training set consists of 1'525 underwater images with annotations for 8 classes. For evaluation, we used a separate split of 110 additional images. The classification labels were estimated from the segmentation ground-truth as a multi-label problem by setting the class label to~1 if the segmentation map contained at least one pixel assigned to that class.
    \item[Cityscapes~\cite{cordts2016cityscapes}:] 2'975 annotated images for both classification and segmentation are available for training. We test on the official Cityscapes validation set, which contains 500 images. 
    \item[OCT:] 22'723 Optical Coherence Tomography~(OCT) images with classification annotations and 1,002~images with pixel-wise annotations corresponding to 4 different types of retinal fluid for segmentation. We split the data into 902~training images and 100~test images.
\end{description}

\subsection{Baseline strategies.} We compared our method to ten different {\it fixed} budget allocation strategies. Each of these randomly sample images for classification and segmentation annotations according to a specified and fixed proportion. We denote these policies by the percentage dedicated to segmentation annotations: $B_0$: $50\%, 55\%, \ldots, 95\%$ with increases in 5\%. For fair comparison, the strategies are computed from the budget $B_0$.

In addition, we consider an {\it estimated-best-fixed} budget allocation strategy, whereby the method estimates what fixed budget should be used for a given dataset. This is done by using the initial budget $B_0$ to compute the best performing fixed strategy (mentioned above) and then using this fixed strategy for the annotation campaign until budget $B$ is reached. This strategy represents an individual that chooses to explore all fixed strategies for an initial small budget and then exploit it. 

\subsection{Implementation details.} 
{\bf Weakly supervised segmentation model:} To train a segmentation model that uses both segmentation and classifications, we first train the models with the weakly-annotated data~$\T_c$ until convergence and then with the segmentation data~$\T_s$. We use the U-Net segmentation model~\cite{ronneberger2015u} for  OCT, and the DeepLabv3 model~\cite{chen2017rethinking} with a ResNet50 backbone on the SUIM, PASCAL, and Cityscapes. For the U-Net, a classification head is appended at the end of the encoding module for the classification task. For the DeepLab-like models, we train the entire backbone on the classification task and then add the ASPP head for segmentation. In all cases, we use the cross-entropy loss for classification and the average of the Dice loss and the cross-Entropy loss for segmentation. While we choose this training strategy for its simplicity, other cross-task or weakly supervised alternatives could have been used as well~\cite{ahn2018learning,Papandreou15}. Additional details are provided in the supplementary materials.

Note that all models are randomly initialized to maximize the impact of classification labels, as Imagenet-pretraining shares a high resemblance to images in PASCAL and Cityscapes. Failing to do so would lead to classification training not adding significant information and may even hurt performance due to catastrophic forgetting~\cite{mccloskey1989catastrophic}.

{\bf Hyperparameters: }
We measured costs in terms of class-label equivalents setting~$\alpha_c=1$ and leaving only $\alpha_s$ as a hyperparameter of our method. We set~$\alpha_s=12$ for all datasets following previous studies on crowdsourced annotations~\cite{Bearman16}. We predict the first GP surface with 8\% of the dataset for both classification and segmentation. This quantity is reduced for OCT classification and VOC segmentation due to the high number of labels available. In all cases, we fixed the number of iterative steps to 8 and set the learning rate of the GP to 0.1.


\section{Results}
\label{sec:results}
\begin{figure}[h]
\centering
\includegraphics[width=\linewidth]{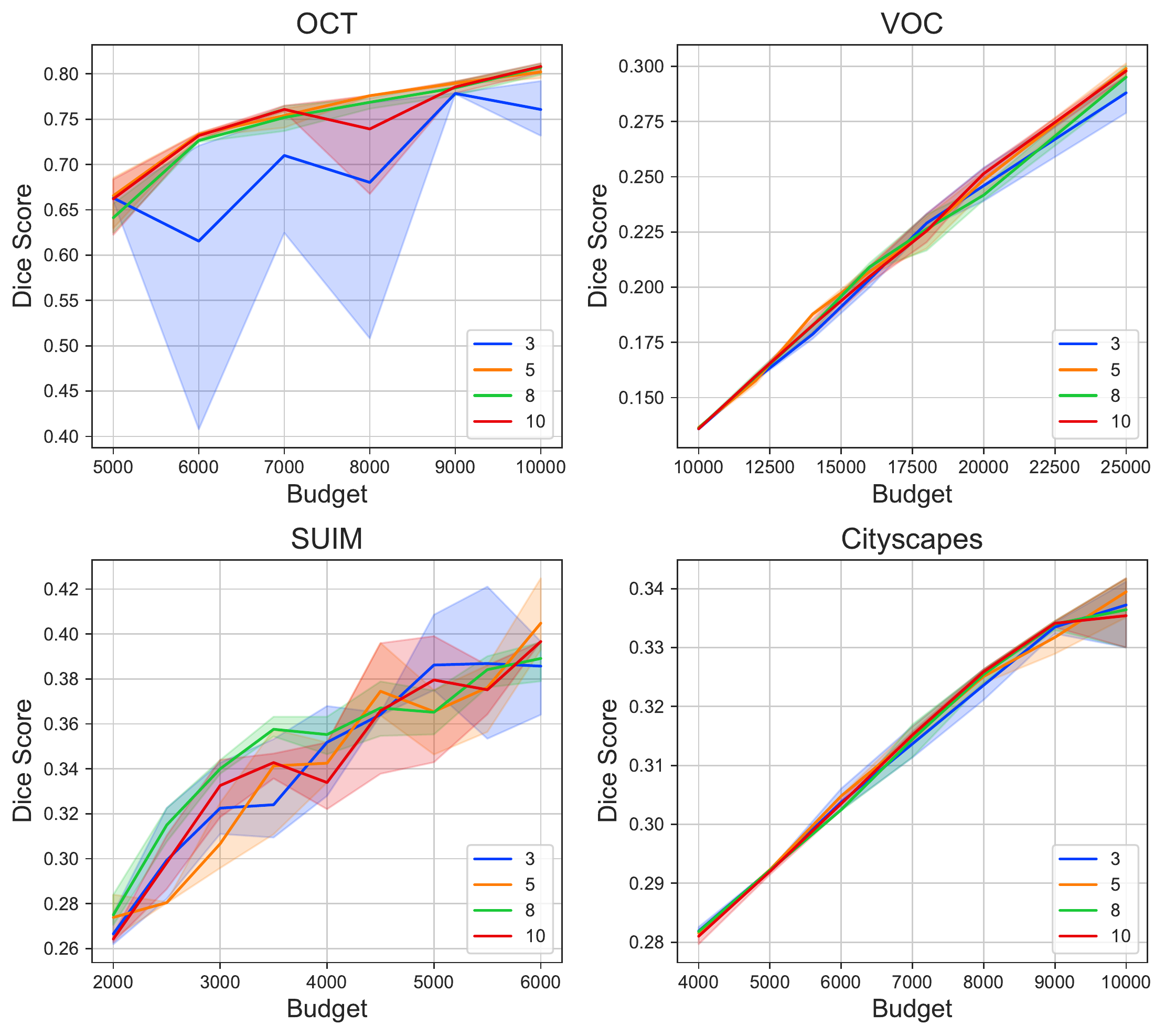}
\caption{Mean of our method when using different numbers of iteration steps $\{3, 5, 8, 10\}$. Results shown with three seeds.
}
\label{fig:steps_results}
\end{figure}
\begin{figure}[t]
\centering
\includegraphics[width=0.71\linewidth]{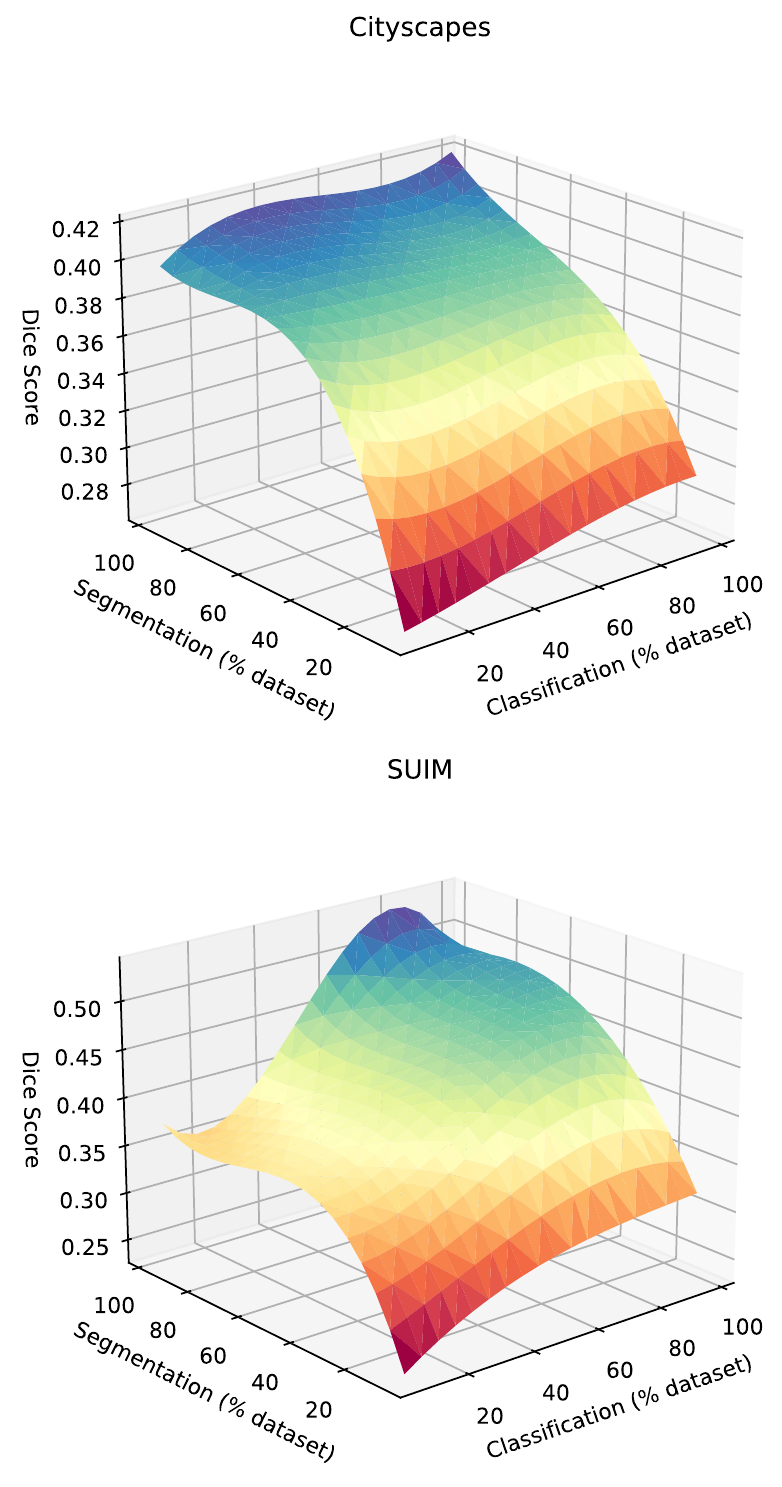}
\caption{Cityscapes (top) and SUIM (bottom) ground truth budget-segmentation surfaces. We note that segmentation performance grows logarithmically with training set size on Cityscapes (as well as OCT and VOC, see the Supplementary materials). This trend is not observed on the SUIM dataset. 
}
\label{fig:surfaces}
\end{figure}
{\bf Main results:} Figure~\ref{fig:main_results} compares the performance achieved by our method against that of the different fixed strategies and the estimated best fixed strategy when using $\alpha_s = 12$ across the different datasets. From these results we can make a number of key observations. 

First, we can observe that no single fixed strategy is performing optimally across the different datasets evaluated. This is coherent with our initial claims and with the literature. Indeed, for OCT the best strategy appears to be one that samples 90\% of segmentations, while this same policy performs poorly on the SUIM dataset. This implies that blindly using a fixed policy would on average not be very effective.

Second, the estimated best-fixed strategy (in red) appears to do well initially and progressively loses competitiveness as the budget increases. This behaviour is expected as the estimated fixed strategy is that with $B_0$ (the lowest budget), and becomes increasingly irrelevant as $B$ grows. This is particularly clear on VOC where the best low-budget strategy allocates 95\% of the budget to segmentation and still achieves superior performance up to $B=12'000$. However, that strategy drops below average performance with budgets greater than $B=25'000$. In the case of SUIM, the best-fixed strategy corresponds to 50\% of the budget allocated to segmentation. Since the dataset contains only 1,525 segmentation samples, this strategy is not attainable with $B > 4000$.

Last, we can observe that our method is consistently able to produce good performances, across both different budget quantities and datasets. We can also clearly see that our strategy is not guaranteed to be the top performing strategy, but that on average it performs well in different cases. 

At the same time, we notice that the performance of our approach on SUIM begins well and then drops after a 3'500 budget. This differs sharply from the other datasets. By observing the true budget-performance surface of SUIM and the other datasets (see \cref{fig:surfaces}), we can see that the SUIM surface does not grow logarithmically with the dataset size, while it does for Cityscapes (and the other too, see the Supplementary materials). This is relevant as our GP mean prior~\eqref{eq:gp_mean} assumes this relationship and explains why our approach fails when the true surface deviates from our GP mean form. While the use of adaptive, higher-level order priors would be beneficial to deal with such cases, we leave this as future work to be researched.


\subsection{Sensitivity to $\alpha_s$ and $T$}
Different types of annotations or domains may have different ratios of cost. While we have fixed $\alpha_s$ in our experiments across all datasets regardless of their domain, some datasets such as OCT and VOC require different expertise and domain knowledge to annotate and thus different $\alpha_s$. In~Fig.~\ref{fig:alpha_results}, we three additional values of $\alpha_s = \{5, 12, 25, 50\}$ and show the performance implication it has on our methods and the baselines. For Cityscapes, we see that the method is robust regardless of the value of $\alpha_s$, showing above average performance especially for $\alpha_s = 25$ and $\alpha_s = 50$. This behavior is reproduced in all four datasets (see Supplementary materials).

Similarly, the number of steps $T$ given to reach the final budget is a hyperparameter of our approach. While low $T$ values could lead to poor solutions due to the unreliability of the GP far from the sampled region, higher $T$ values (\ie, therefore smaller steps) may exacerbate the intrinsic greedy nature of our method. We thus seek a trade-off between reliability and greediness. To study the sensitivity of the algorithm with respect to this variable, we show the behaviour of our method with different number of steps in Fig.~\ref{fig:steps_results}. We see that lower $T$ values greatly affect the reliability of the found strategy, especially for OCT and SUIM (blue line). However, as the number of steps increases, the variance of the strategy reduces sharply. We can therefore conclude that the method is robust to this hyperparameter as long as it is kept within reasonable ranges.

\section{Conclusion}
\label{sec:conclusion}
In this paper, we propose a novel approach to determine a dynamic annotation strategy for building segmentation datasets. We design an iterative process that identifies efficient dataset-specific combinations of weak annotations in the form of image-level labels and full segmentations. We show in our experiments that the best strategies are often dataset and budget-dependent, and therefore the trivial approaches do not always produce the best results. Our method however is capable of adapting to different image domains and finds combinations of annotations that reach high-performance levels. We show our method is robust to a number of hyperparameters and that it offers a good option for allocating annotation strategies.

\newpage

{\small
\bibliographystyle{ieee_fullname}
\bibliography{00_main}

\begin{thebibliography}{10}\itemsep=-1pt

\bibitem{ahn2019weakly}
Jiwoon Ahn, Sunghyun Cho, and Suha Kwak.
\newblock {Weakly supervised learning of instance segmentation with inter-pixel
  relations}.
\newblock In {\em Proceedings of the IEEE/CVF conference on computer vision and
  pattern recognition}, pages 2209--2218, 2019.

\bibitem{ahn2018learning}
Jiwoon Ahn and Suha Kwak.
\newblock {Learning pixel-level semantic affinity with image-level supervision
  for weakly supervised semantic segmentation}.
\newblock In {\em Proceedings of the IEEE conference on computer vision and
  pattern recognition}, pages 4981--4990, 2018.

\bibitem{andriluka2018fluid}
Mykhaylo Andriluka, Jasper R~R Uijlings, and Vittorio Ferrari.
\newblock {Fluid annotation: a human-machine collaboration interface for full
  image annotation}.
\newblock In {\em Proceedings of the 26th ACM international conference on
  Multimedia}, pages 1957--1966, 2018.

\bibitem{Bearman16}
Amy Bearman, Olga Russakovsky, Vittorio Ferrari, and Li Fei-Fei.
\newblock What's the point: Semantic segmentation with point supervision.
\newblock In Bastian Leibe, Jiri Matas, Nicu Sebe, and Max Welling, editors,
  {\em Computer Vision -- ECCV 2016}, pages 549--565, Cham, 2016. Springer
  International Publishing.

\bibitem{Benenson2019LargeScaleIO}
Rodrigo Benenson, Stefan Popov, and Vittorio Ferrari.
\newblock {Large-Scale Interactive Object Segmentation With Human Annotators}.
\newblock {\em 2019 IEEE/CVF Conference on Computer Vision and Pattern
  Recognition (CVPR)}, pages 11692--11701, 2019.

\bibitem{Bodenstedt2018}
Sebastian Bodenstedt, Max Allan, Anthony Agustinos, Xiaofei Du, Luis~C.
  Garc{\'{\i}}a{-}Peraza{-}Herrera, Hannes Kenngott, Thomas Kurmann, Beat~P.
  M{\"{u}}ller{-}Stich, S{\'{e}}bastien Ourselin, Daniil Pakhomov, Raphael
  Sznitman, Marvin Teichmann, Martin Thoma, Tom Vercauteren, Sandrine Voros,
  Martin Wagner, Pamela Wochner, Lena Maier{-}Hein, Danail Stoyanov, and
  Stefanie Speidel.
\newblock Comparative evaluation of instrument segmentation and tracking
  methods in minimally invasive surgery.
\newblock {\em ArXiv}, abs/1805.02475, 2018.

\bibitem{Cai21}
Lile Cai, Xun Xu, Jun~Hao Liew, and Chuan Sheng~Foo.
\newblock Revisiting superpixels for active learning in semantic segmentation
  with realistic annotation costs.
\newblock In {\em 2021 IEEE/CVF Conference on Computer Vision and Pattern
  Recognition (CVPR)}, pages 10983--10992, 2021.

\bibitem{Casanova2020Reinforced}
Arantxa Casanova, Pedro~O. Pinheiro, Negar Rostamzadeh, and Christopher~J. Pal.
\newblock Reinforced active learning for image segmentation.
\newblock In {\em International Conference on Learning Representations}, 2020.

\bibitem{chen2020deep}
Chen Chen, Chen Qin, Huaqi Qiu, Giacomo Tarroni, Jinming Duan, Wenjia Bai, and
  Daniel Rueckert.
\newblock {Deep learning for cardiac image segmentation: a review}.
\newblock {\em Frontiers in Cardiovascular Medicine}, page~25, 2020.

\bibitem{chen2017rethinking}
Liang-Chieh Chen, George Papandreou, Florian Schroff, and Hartwig Adam.
\newblock {Rethinking atrous convolution for semantic image segmentation}.
\newblock {\em arXiv preprint arXiv:1706.05587}, 2017.

\bibitem{cordts2016cityscapes}
Marius Cordts, Mohamed Omran, Sebastian Ramos, Timo Rehfeld, Markus Enzweiler,
  Rodrigo Benenson, Uwe Franke, Stefan Roth, and Bernt Schiele.
\newblock {The cityscapes dataset for semantic urban scene understanding}.
\newblock In {\em Proceedings of the IEEE conference on computer vision and
  pattern recognition}, pages 3213--3223, 2016.

\bibitem{deng2009imagenet}
Jia Deng, Wei Dong, Richard Socher, Li-Jia Li, Kai Li, and Li Fei-Fei.
\newblock {Imagenet: A large-scale hierarchical image database}.
\newblock In {\em 2009 IEEE conference on computer vision and pattern
  recognition}, pages 248--255. Ieee, 2009.

\bibitem{Ding2019}
Yiming Ding, Jae~Ho Sohn, Michael~G Kawczynski, Hari Trivedi, Roy Harnish,
  Nathaniel~W Jenkins, Dmytro Lituiev, Timothy~P Copeland, Mariam~S Aboian,
  Carina {Mari Aparici}, Spencer~C Behr, Robert~R Flavell, Shih-Ying Huang,
  Kelly~A Zalocusky, Lorenzo Nardo, Youngho Seo, Randall~A Hawkins, Miguel
  {Hernandez Pampaloni}, Dexter Hadley, and Benjamin~L Franc.
\newblock {A Deep Learning Model to Predict a Diagnosis of Alzheimer Disease by
  Using 18F-FDG PET of the Brain}.
\newblock {\em Radiology}, 290(2):456--464, 2019.

\bibitem{Esteva2017}
Andre Esteva, Brett Kuprel, Roberto~A Novoa, Justin Ko, Susan~M Swetter,
  Helen~M Blau, and Sebastian Thrun.
\newblock {Dermatologist-level classification of skin cancer with deep neural
  networks}.
\newblock {\em Nature}, 542(7639):115--118, 2017.

\bibitem{pascal-voc-2012}
M Everingham, L Van{\~{}}Gool, C~K~I Williams, J Winn, and A Zisserman.
\newblock {The PASCAL Visual Object Classes Challenge 2012 VOC2012 Results}.
\newblock
  http://www.pascal-network.org/challenges/VOC/voc2012/workshop/index.html.

\bibitem{fan2020learning}
Junsong Fan, Zhaoxiang Zhang, Chunfeng Song, and Tieniu Tan.
\newblock {Learning integral objects with intra-class discriminator for
  weakly-supervised semantic segmentation}.
\newblock In {\em Proceedings of the IEEE/CVF Conference on Computer Vision and
  Pattern Recognition}, pages 4283--4292, 2020.

\bibitem{heker2020joint}
Michal Heker and Hayit Greenspan.
\newblock {Joint liver lesion segmentation and classification via transfer
  learning}.
\newblock {\em arXiv preprint arXiv:2004.12352}, 2020.

\bibitem{hesamian2019deep}
Mohammad~Hesam Hesamian, Wenjing Jia, Xiangjian He, and Paul Kennedy.
\newblock {Deep learning techniques for medical image segmentation:
  achievements and challenges}.
\newblock {\em Journal of digital imaging}, 32(4):582--596, 2019.

\bibitem{islam2020semantic}
Md~Jahidul Islam, Chelsey Edge, Yuyang Xiao, Peigen Luo, Muntaqim Mehtaz,
  Christopher Morse, Sadman~Sakib Enan, and Junaed Sattar.
\newblock {Semantic segmentation of underwater imagery: Dataset and benchmark}.
\newblock In {\em 2020 IEEE/RSJ International Conference on Intelligent Robots
  and Systems (IROS)}, pages 1769--1776. IEEE, 2020.

\bibitem{jiang2019integral}
Peng-Tao Jiang, Qibin Hou, Yang Cao, Ming-Ming Cheng, Yunchao Wei, and Hong-Kai
  Xiong.
\newblock {Integral object mining via online attention accumulation}.
\newblock In {\em Proceedings of the IEEE/CVF International Conference on
  Computer Vision}, pages 2070--2079, 2019.

\bibitem{jones1998efficient}
Donald~R Jones, Matthias Schonlau, and William~J Welch.
\newblock {Efficient global optimization of expensive black-box functions}.
\newblock {\em Journal of Global optimization}, 13(4):455--492, 1998.

\bibitem{Joshi2009}
Ajay~J Joshi, Fatih Porikli, and Nikolaos Papanikolopoulos.
\newblock {Multi-class active learning for image classification}.
\newblock In {\em 2009 IEEE Conference on Computer Vision and Pattern
  Recognition}, pages 2372--2379, 2009.

\bibitem{Ki2021}
Minsong Ki, Youngjung Uh, Wonyoung Lee, and Hyeran Byun.
\newblock {Contrastive and consistent feature learning for weakly supervised
  object localization and semantic segmentation}.
\newblock {\em Neurocomputing}, 445:244--254, jul 2021.

\bibitem{kolesnikov2020big}
Alexander Kolesnikov, Lucas Beyer, Xiaohua Zhai, Joan Puigcerver, Jessica Yung,
  Sylvain Gelly, and Neil Houlsby.
\newblock {Big transfer (bit): General visual representation learning}.
\newblock In {\em European conference on computer vision}, pages 491--507.
  Springer, 2020.

\bibitem{koleshnikov2021}
Alexander Kolesnikov, Alexey Dosovitskiy, Dirk Weissenborn, Georg Heigold,
  Jakob Uszkoreit, Lucas Beyer, Matthias Minderer, Mostafa Dehghani, Neil
  Houlsby, Sylvain Gelly, Thomas Unterthiner, and Xiaohua Zhai.
\newblock {An Image is Worth 16x16 Words: Transformers for Image Recognition at
  Scale}.
\newblock 2021.

\bibitem{Konyushkova15}
Ksenia Konyushkova, Raphael Sznitman, and Pascal Fua.
\newblock Introducing geometry in active learning for image segmentation.
\newblock In {\em 2015 IEEE International Conference on Computer Vision
  (ICCV)}, pages 2974--2982, 2015.

\bibitem{Konyushkova_2018_CVPR}
Ksenia Konyushkova, Jasper Uijlings, Christoph~H Lampert, and Vittorio Ferrari.
\newblock {Learning Intelligent Dialogs for Bounding Box Annotation}.
\newblock In {\em Proceedings of the IEEE Conference on Computer Vision and
  Pattern Recognition (CVPR)}, jun 2018.

\bibitem{lee2019ficklenet}
Jungbeom Lee, Eunji Kim, Sungmin Lee, Jangho Lee, and Sungroh Yoon.
\newblock {Ficklenet: Weakly and semi-supervised semantic image segmentation
  using stochastic inference}.
\newblock In {\em Proceedings of the IEEE/CVF Conference on Computer Vision and
  Pattern Recognition}, pages 5267--5276, 2019.

\bibitem{li2020mas3k}
Lin Li, Eric Rigall, Junyu Dong, and Geng Chen.
\newblock {MAS3K: An Open Dataset for Marine Animal Segmentation}.
\newblock In {\em International Symposium on Benchmarking, Measuring and
  Optimization}, pages 194--212. Springer, 2020.

\bibitem{Liang2020}
Gaobo Liang and Lixin Zheng.
\newblock {A transfer learning method with deep residual network for pediatric
  pneumonia diagnosis}.
\newblock {\em Computer Methods and Programs in Biomedicine}, 187:104964, apr
  2020.

\bibitem{lin2016scribblesup}
Di Lin, Jifeng Dai, Jiaya Jia, Kaiming He, and Jian Sun.
\newblock {Scribblesup: Scribble-supervised convolutional networks for semantic
  segmentation}.
\newblock In {\em Proceedings of the IEEE conference on computer vision and
  pattern recognition}, pages 3159--3167, 2016.

\bibitem{liu2020fsd}
Shenlan Liu, Xiang Liu, Gao Huang, Lin Feng, Lianyu Hu, Dong Jiang, Aibin
  Zhang, Yang Liu, and Hong Qiao.
\newblock {FSD-10: a dataset for competitive sports content analysis}.
\newblock {\em arXiv preprint arXiv:2002.03312}, 2020.

\bibitem{mahmood2022}
Rafid Mahmood, James Lucas, David Acuna, Daiqing Li, Jonah Philion, Jose~M
  Alvarez, Zhiding Yu, Sanja Fidler, and Marc~T Law.
\newblock How much more data do i need? estimating requirements for downstream
  tasks.
\newblock {\em Proceedings of the IEEE/CVF Conference on Computer Vision and
  Pattern Recognition}, pages 275--284, 2022.

\bibitem{mahmood2022optimizing}
Rafid Mahmood, James Lucas, Jose~M. Alvarez, Sanja Fidler, and Marc~T. Law.
\newblock Optimizing data collection for machine learning.
\newblock {\em Advances in Neural Information Processing Systems (NeurIPS)}, 10
  2022.

\bibitem{mccloskey1989catastrophic}
Michael McCloskey and Neal~J Cohen.
\newblock {Catastrophic interference in connectionist networks: The sequential
  learning problem}.
\newblock In {\em Psychology of learning and motivation}, volume~24, pages
  109--165. Elsevier, 1989.

\bibitem{menegola2017}
Afonso Menegola, Michel Fornaciali, Ramon Pires, Fl{\'{a}}via~Vasques
  Bittencourt, Sandra Avila, and Eduardo Valle.
\newblock {Knowledge transfer for melanoma screening with deep learning}.
\newblock In {\em 2017 IEEE 14th International Symposium on Biomedical Imaging
  (ISBI 2017)}, pages 297--300. IEEE, 2017.

\bibitem{Mensink}
Thomas Mensink, Jasper Uijlings, Alina Kuznetsova, Michael Gygli, and Vittorio
  Ferrari.
\newblock {Factors of Influence for Transfer Learning across Diverse Appearance
  Domains and Task Types}.
\newblock {\em arXiv preprint arXiv:2103.13318}, 2021.

\bibitem{papadopoulos2014}
Dim~P Papadopoulos, Alasdair D~F Clarke, Frank Keller, and Vittorio Ferrari.
\newblock {Training Object Class Detectors from Eye Tracking Data}.
\newblock In David Fleet, Tomas Pajdla, Bernt Schiele, and Tinne Tuytelaars,
  editors, {\em Computer Vision -- ECCV 2014}, pages 361--376, Cham, 2014.
  Springer International Publishing.

\bibitem{papadopoulos2017extreme}
Dim~P Papadopoulos, Jasper R~R Uijlings, Frank Keller, and Vittorio Ferrari.
\newblock {Extreme clicking for efficient object annotation}.
\newblock In {\em Proceedings of the IEEE international conference on computer
  vision}, pages 4930--4939, 2017.

\bibitem{Papandreou15}
George Papandreou, Liang-Chieh Chen, Kevin~P. Murphy, and Alan~L. Yuille.
\newblock Weakly- and semi-supervised learning of a deep convolutional network
  for semantic image segmentation.
\newblock In {\em Proceedings of the IEEE International Conference on Computer
  Vision (ICCV)}, December 2015.

\bibitem{Ranganathan2017}
Hiranmayi Ranganathan, Hemanth Venkateswara, Shayok Chakraborty, and Sethuraman
  Panchanathan.
\newblock {Deep active learning for image classification}.
\newblock In {\em 2017 IEEE International Conference on Image Processing
  (ICIP)}, pages 3934--3938, 2017.

\bibitem{ronneberger2015u}
Olaf Ronneberger, Philipp Fischer, and Thomas Brox.
\newblock {U-net: Convolutional networks for biomedical image segmentation}.
\newblock In {\em International Conference on Medical image computing and
  computer-assisted intervention}, pages 234--241. Springer, 2015.

\bibitem{siam2017deep}
Mennatullah Siam, Sara Elkerdawy, Martin Jagersand, and Senthil Yogamani.
\newblock {Deep semantic segmentation for automated driving: Taxonomy, roadmap
  and challenges}.
\newblock In {\em 2017 IEEE 20th international conference on intelligent
  transportation systems (ITSC)}, pages 1--8. IEEE, 2017.

\bibitem{siddiqui2020viewal}
Yawar Siddiqui, Julien Valentin, and Matthias Nie{\ss}ner.
\newblock {Viewal: Active learning with viewpoint entropy for semantic
  segmentation}.
\newblock In {\em Proceedings of the IEEE/CVF conference on computer vision and
  pattern recognition}, pages 9433--9443, 2020.

\bibitem{sless2019road}
Liat Sless, Bat {El Shlomo}, Gilad Cohen, and Shaul Oron.
\newblock {Road scene understanding by occupancy grid learning from sparse
  radar clusters using semantic segmentation}.
\newblock In {\em Proceedings of the IEEE/CVF International Conference on
  Computer Vision Workshops}, page~0, 2019.

\bibitem{song2019box}
Chunfeng Song, Yan Huang, Wanli Ouyang, and Liang Wang.
\newblock {Box-driven class-wise region masking and filling rate guided loss
  for weakly supervised semantic segmentation}.
\newblock In {\em Proceedings of the IEEE/CVF Conference on Computer Vision and
  Pattern Recognition}, pages 3136--3145, 2019.

\bibitem{sun2017}
Chen Sun, Abhinav Shrivastava, Saurabh Singh, and Abhinav Gupta.
\newblock {Revisiting unreasonable effectiveness of data in deep learning era}.
\newblock In {\em Proceedings of the IEEE international conference on computer
  vision}, pages 843--852, 2017.

\bibitem{Tajbakhsh2016}
Nima Tajbakhsh, Jae~Y. Shin, Suryakanth~R. Gurudu, R.~Todd Hurst,
  Christopher~B. Kendall, Michael~B. Gotway, and Jianming Liang.
\newblock {Convolutional Neural Networks for Medical Image Analysis: Full
  Training or Fine Tuning?}
\newblock {\em IEEE Transactions on Medical Imaging}, 35(5):1299--1312, may
  2016.

\bibitem{tang2018normalized}
Meng Tang, Abdelaziz Djelouah, Federico Perazzi, Yuri Boykov, and Christopher
  Schroers.
\newblock {Normalized cut loss for weakly-supervised cnn segmentation}.
\newblock In {\em Proceedings of the IEEE conference on computer vision and
  pattern recognition}, pages 1818--1827, 2018.

\bibitem{tseng2021person}
Chien-Hao Tseng, Chia-Chien Hsieh, Dah-Jing Jwo, Jyh-Horng Wu, Ruey-Kai Sheu,
  and Lun-Chi Chen.
\newblock {Person Retrieval in Video Surveillance Using Deep Learning--Based
  Instance Segmentation}.
\newblock {\em Journal of Sensors}, 2021, 2021.

\bibitem{wang2017chestxray}
Xiaosong Wang, Yifan Peng, Le Lu, Zhiyong Lu, Mohammadhadi Bagheri, and Ronald
  Summers.
\newblock {ChestX-ray8: Hospital-scale Chest X-ray Database and Benchmarks on
  Weakly-Supervised Classification and Localization of Common Thorax Diseases}.
\newblock In {\em 2017 IEEE Conference on Computer Vision and Pattern
  Recognition(CVPR)}, pages 3462--3471, 2017.

\bibitem{WelinderEtal2010}
P Welinder, S Branson, T Mita, C Wah, F Schroff, S Belongie, and P Perona.
\newblock {Caltech-UCSD Birds 200}.
\newblock Technical Report CNS-TR-2010-001, California Institute of Technology,
  2010.

\bibitem{Zamir2018}
Amir~R Zamir, Alexander Sax, William Shen, Leonidas Guibas, Jitendra Malik, and
  Silvio Savarese.
\newblock {Taskonomy: Disentangling Task Transfer Learning}.
\newblock In {\em Proceedings of the IEEE Conference on Computer Vision and
  Pattern Recognition (CVPR)}, pages 3712--3722, jun 2018.

\bibitem{Zhang2021}
Bingfeng Zhang, Jimin Xiao, Jianbo Jiao, Yunchao Wei, and Yao Zhao.
\newblock {Affinity Attention Graph Neural Network for Weakly Supervised
  Semantic Segmentation}.
\newblock {\em IEEE Transactions on Pattern Analysis and Machine Intelligence},
  pages 1--1, 2021.

\bibitem{zhou2016learning}
Bolei Zhou, Aditya Khosla, Agata Lapedriza, Aude Oliva, and Antonio Torralba.
\newblock {Learning deep features for discriminative localization}.
\newblock In {\em Proceedings of the IEEE conference on computer vision and
  pattern recognition}, pages 2921--2929, 2016.

\end{thebibliography}
}

\newpage
\onecolumn
\appendix
\section{Implementation details}
\label{sec:impl}
{\bf Weakly-supervised segmentation model hyperparameters: }
We used the U-Net segmentation model~\cite{ronneberger2015u} for OCT, and the DeepLabV3 model~\cite{chen2017rethinking} with a ResNet50 backbone on the SUIM, PASCAL, and Cityscapes datasets. For the U-Net, a classification head with max-avg pooling and one fully connected layer was appended at the end of the encoding module for the classification task. For the DeepLab-like models, we trained the entire ResNet-50 backbone on the classification task and then added the ASPP head for segmentation. In all cases, we used the cross-entropy loss for classification and the average of the Dice and cross-entropy losses for segmentation. \cref{tab:configurations_model} contains the details of batch sizes and optimizers.

\begin{table}[h]
\centering
\begin{tabular}{lcccccc}
\toprule
\textbf{Dataset} & \textbf{Model} &  \textbf{Optimizer} & \textbf{Batch size} \\ \midrule
\textbf{OCT} & U-Net &  Adam & 8  \\
\textbf{VOC} & DeepLabV3 &  SGD & 16 \\
\textbf{SUIM} & DeepLabV3 &  SGD & 8  \\ 
\textbf{Cityscapes} & DeepLabV3 & SGD & 16   \\ \bottomrule
\end{tabular}
\caption{Hyperparameters and conditions for all experiments.}
\label{tab:configurations_model}
\end{table}

{\bf Algorithm hyperparameters: }
We measured the costs of annotations in terms of class-label equivalents setting~$\alpha_c=1$ and leaving only $\alpha_s$ as a hyperparameter of our method. We set to~$\alpha_s=12$ for all datasets following previous studies on crowdsourced annotations~\cite{Bearman16}. We fixed the number of iterative steps to~$T=8$ and the learning rate of the GP to~$0.1$. We set both the initial number of class annotations~$C_0$ and segmentation annotations~$S_0$ to~$8\%$ of the available labels for SUIM and Cityscapes. We reduced $C_0$~in OCT and $S_0$~in VOC to account for the higher number of labels available in those datasets, as detailed in~\cref{tab:configurations_gp}.

\begin{table}[h]
\centering
\begin{tabular}{lcccccc}
\toprule
\textbf{Dataset} & $C_0 (\%)$ & $S_0 (\%)$ & $B_0$ \\ \midrule
\textbf{OCT} & 4 & 8 & 1'774 \\
\textbf{VOC} & 8 & 6 & 8'076 \\
\textbf{SUIM} & 8 & 8 & 1'586 \\ 
\textbf{Cityscapes} & 8 & 8 & 1'774  \\ \bottomrule
\end{tabular}
\caption{Initial conditions for our method. $B_0$ calculated with $\alpha_s = 12$ and $\alpha_c = 1$.}
\label{tab:configurations_gp}
\end{table}

\section{Sensitivity to $\alpha_s$}
\label{sec:sensitivity}

\begin{figure}[h]
\centering
\includegraphics[width=\linewidth]{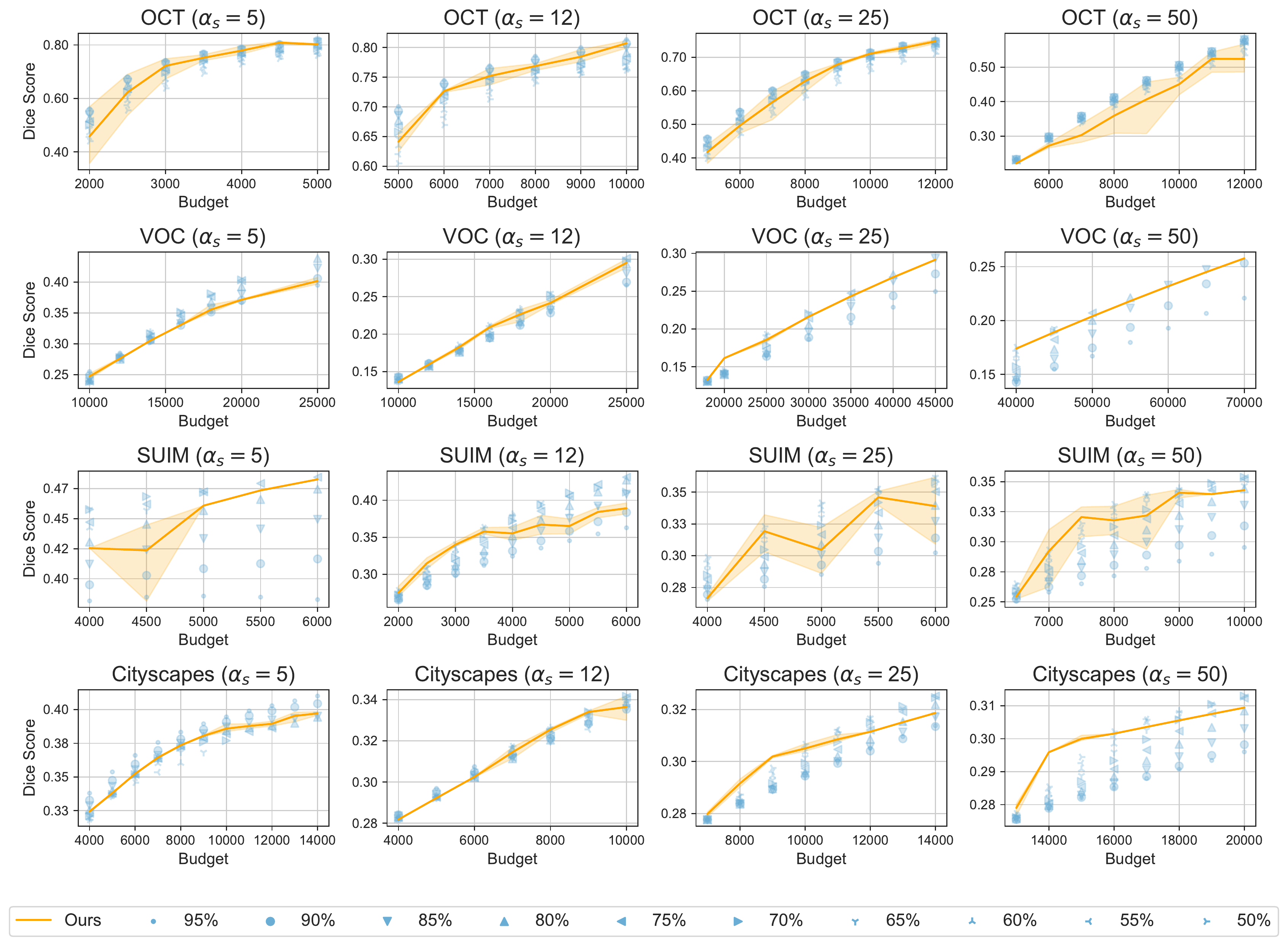}
\caption{Performance our method with $\alpha_s = \{5, 12, 25, 50\}$ on four datasets. One-sigma error bars were computed from three seeds. Blue marks show the performance of fixed strategies, with labels indicating the percentage of the budget allocated to segmentation annotations.
}
\label{fig:alpha_results_supp}
\end{figure}

\cref{fig:alpha_results_supp} shows aditional experiments on the sensitivity of our method to~$\alpha_s$ in the considered datasets.

\section{Average performance}
\label{sec:average}

We report in Table~\ref{tab:average_performance} the average relative performance for each dataset and segmentation split. Our method is best on average and for two datasets (SUIM and VOC), while it performs on par with the best baseline for Cityscapes and OCT.

\begin{table}[htbp]
\centering
\begin{tabular}{@{}cccccccccccc@{}}
\toprule
Seg. Split & 95 & 90 & 85 & 80 & 75 & 70 & 65 & 60 & 55 & 50 & Ours \\ \midrule
Cityscapes & \textbf{73.9} & 73.2 & 73.0 & 73.1 & 73.3 & 73.5 & 73.5 & 72.5 & 72.3 & 71.4 & 73.4 \\
OCT & \textbf{95.4} & 94.6 & 93.8 & 93.0 & 92.1 & 91.3 & 90.5 & 89.7 & 88.8 & 87.7 & 92.6 \\
SUIM & 72.1 & 74.1 & 76.7 & 79.2 & 80.9 & 81.3 & 79.7 & 78.1 & 76.7 & 75.1 & \textbf{81.9} \\
VOC & 44.3 & 44.0 & 44.3 & 44.9 & 45.6 & 46.0 & 43.1 & 43.3 & 40.9 & 40.8 & \textbf{46.0} \\ \hline
Average & 71.4 & 71.5 & 71.9 & 72.5 & 73.0 & 73.0 & 71.7 & 70.9 & 69.7 & 68.8 & \textbf{73.5} \\ \bottomrule
\end{tabular}
\caption{Relative performance (in \%) against full supervision for each dataset and segmentation split.}
\label{tab:average_performance}
\end{table}
\section{Surfaces}
\label{sec:surfaces}

\begin{figure}[h]
\centering
\includegraphics[width=0.8\linewidth]{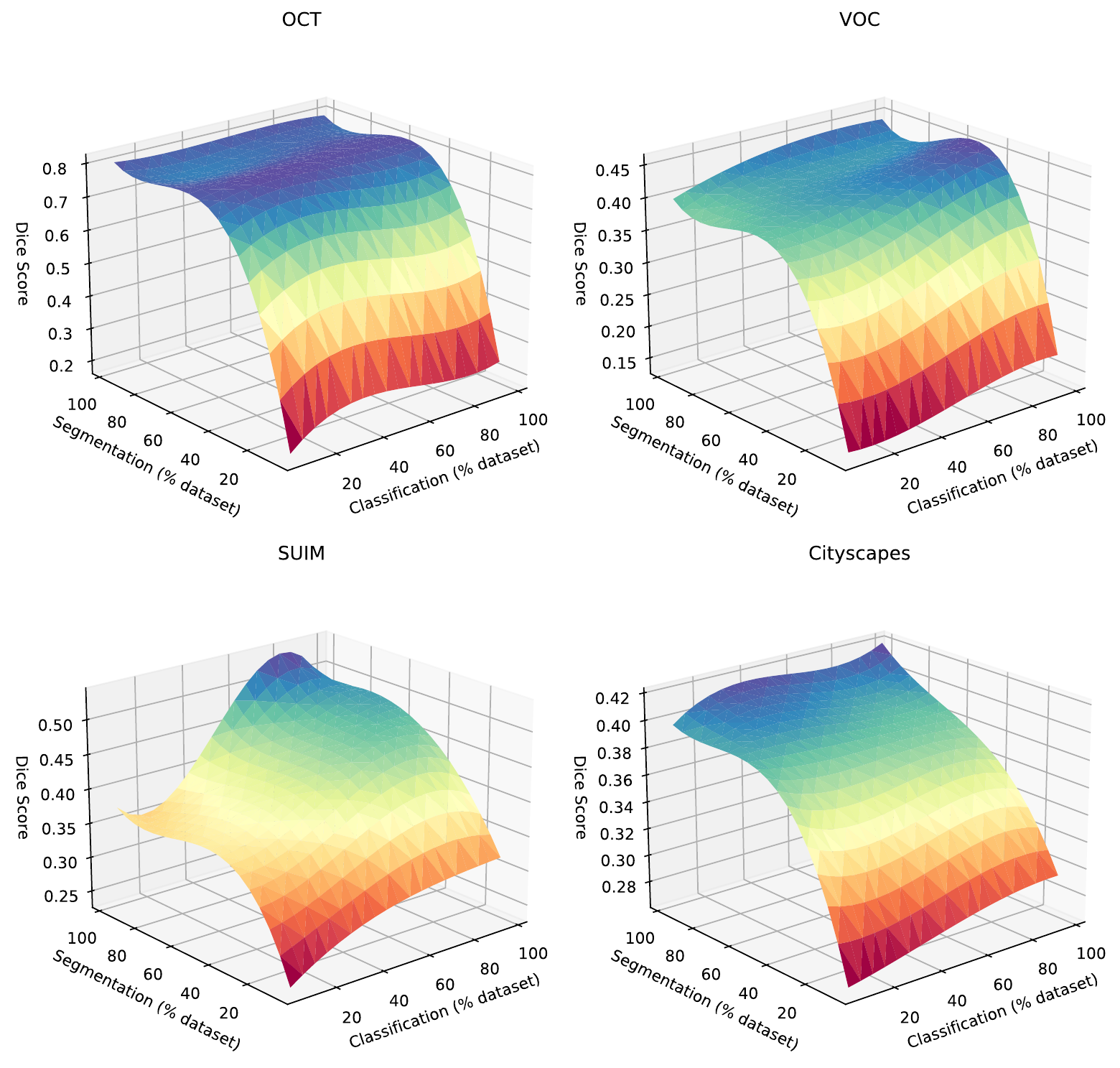}
\caption{Segmentation performance grows logarithmically with training set size on Cityscapes, OCT, and VOC. This trend is not observed in the SUIM dataset. 
}
\label{fig:surface_supp}
\end{figure}

To explain the drop in performance for SUIM after a budget of 3'500, we computed the true surfaces for all datasets~(\cref{fig:surface_supp}). We observed that segmentation performance grows logarithmically for OCT, VOC, and Cityscapes, but not for SUIM after a certain number of class annotations. Since our GP assumes a logarithmic relation between dataset size and performance, this observation is particularly relevant to explain the decline in performance for SUIM. Most notably, and confirming our hypothesis, this decline is not seen with other values of~$\alpha_s$ for this same dataset~(\cref{fig:alpha_results_supp}). Due to the limited dataset size, larger values of $\alpha_s$ constrain the area of the surface that can be reached in our experiments. In the case of SUIM, $\alpha_s = 50$ translates to exploring zones where performance grows logarithmically with dataset size (low data regime).

\section{Ground-truth approximation}
\label{sec:gt}
To address the high computational demands of our experiments, we followed the procedure of~\cite{mahmood2022optimizing} for ground-truth approximation. 
In particular, we built subsets of the training dataset by randomly sampling different proportions of the available annotated samples~\cite{mahmood2022}. The proportions~$\rho_s$ and~$\rho_c$ for segmentation and classification samples, respectively, were chosen from the set~$\{2\%, 4\%,	6\%, 8\%, 10\%, 20\%, 30\%, 40\%, 60\%, 80\%, 100\%\}$, for a total of $11\times{}11=121$~possible training subsets. For each subset, we trained the weakly-supervised segmentation model and measured its Dice score on a fixed segmentation test set. We finally interpolated these scores with third-order splines to obtain a surface of ground-truth Dice scores. This procedure allowed efficient estimations of the Dice Score values without retraining a new model for each strategy.

The proportions~$\rho_s$ and~$\rho_c$ are relative to the total amounts of available annotated samples in each dataset, which are shown in \cref{tab:datasets}.


\begin{table}[h]
\centering
\begin{tabular}{lcc}
\hline
\textbf{Dataset} & \textbf{Segmentation} & \textbf{Classification} \\ \hline
\textbf{OCT} & 902 & 22'723 \\
\textbf{VOC} & 10'582 & 5'717 \\
\textbf{SUIM} & 1'525 & 1'525 \\
\textbf{Cityscapes} & 2'975 & 2'975 \\ \hline
\end{tabular}
\caption{Number of training images for each dataset and modality.}
\label{tab:datasets}
\end{table}

\end{document}